# MATCHING OF SAR AND OPTICAL IMAGES BASED ON TRANSFORMATION TO SHARED MODALITY

**Alexey Borisov, Evgeny Myasnikov, Vladislav Myasnikov**

Samara National Research University
Samara, Russia
borisov.an@ssau.ru, mevg@geosamara.ru, vmyas@geosamara.ru

February 13, 2026

## ABSTRACT

Significant differences in optical images and Synthetic Aperture Radar (SAR) images are caused by fundamental differences in the physical principles underlying their acquisition by Earth remote sensing platforms. These differences make precise image matching (co-registration) of these two types of images difficult. In this paper, we propose a new approach to image matching of optical and SAR images, which is based on transforming the images to a new modality. The new image modality is common to both optical and SAR images and satisfies the following conditions. First, the transformed images must have an equal pre-defined number of channels. Second, the transformed and co-registered images must be as similar as possible. Third, the transformed images must be non-degenerate, meaning they must preserve the significant features of the original images. To further match images transformed to this shared modality, we train the RoMa image matching model, which is one of the leading solutions for matching of regular digital photographs. We evaluated the proposed approach on the publicly available MultiSenGE dataset containing both optical and SAR images. We demonstrated its superiority over alternative approaches based on image translation between original modalities and various feature matching algorithms. The proposed solution not only provides better quality of matching, but is also more versatile. It enables the use of ready-made RoMa and DeDoDe models, pre-trained for regular images, without retraining for a new modality, while maintaining high-quality matching of optical and SAR images.

## 1 Introduction

Earth observation remote sensing tools are actively and successfully used in the modern global economy. Satellites with optical (RGB, multispectral, hyperspectral) and SAR sensors are the most in demand. Their number is constantly growing as more countries and large commercial companies become involved in the Earth observation process. As the volume of Earth observation data grows, the problem of satellite image matching is becoming increasingly important. This problem is particularly relevant when the images are obtained from sensors using different physical principles, the images have different spatial/spectral resolution, or were taken at different points in time.

The problem of digital image matching, also called the co-registration problem, has often arisen in applications of photogrammetry and related areas of technical vision [1, 2]. For instance, finding corresponding regions on different images is a key step for constructing a 3D model of an object [3], solving the problem of simultaneous localization and mapping (SLAM) [4], change detection [5], image fusion [6], etc.

The development of imaging tools of various types (RGB, video, multi- and hyperspectral, etc.), personal mobile devices, and high-performance data processing tools has led to an explosive growth in computer vision solutions. In particular, image co-registration techniques have been intensively developed. As a result, these techniques have come a long way from simple correlation and edge detection methods [7] to dense matching methods using deep neural networks [1, 2].

Despite the significant diversity of image matching methods for conventional RGB images, developing similar methods for images acquired by different sensors (multi- and hyperspectral, SAR) remains challenging due to the significant differences between such images (see Figure 1).

The problem of accurate co-registration is of particular interest for optical and SAR images. On the one hand, optical images can easily be interpreted by humans and contain information at a range of spectral bands (excluding the case of a single-band sensor, e.g., the Cartosat-2B panchromatic sensor), but the quality of optical images depends on the weather. On the other hand, SAR images are available at any time of day and in any weather conditions, but are prone to speckle noise.

In recent years, there has been a trend toward using multimodal and/or fundamental artificial intelligence models [8-10], which simultaneously utilize optical and SAR images to improve the quality of data analysis. However, both SAR and optical images have to be precisely co-registered in order to perform such joint analysis.

Modern deep learning - based methods for co-registration of optical and SAR remote sensing images have received the most attention recently. These methods can be divided into two categories [11].

*The first class* of methods uses deep learning to construct descriptors, which allow for matching optical and SAR images. The examples of this class include MatchosNet [12], HardNet [13], ADRNet [14], etc. Methods of *the second class* transform images obtained from one sensor (the first modality) to a representation similar to images registered by the second sensor (the second modality). Such a translation between modalities, reminiscent of image style transfer [15], is typical for generative artificial intelligence tools [16, 17]. Solutions of the second class, based on generative adversarial networks (GANs), were used in [18, 19] to match optical and SAR images. Similar approaches were also applied to images of other modalities [20, 21].

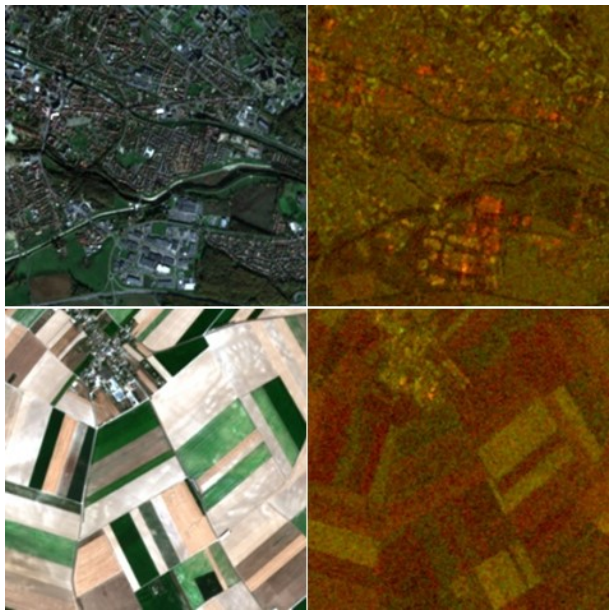

Figure 1: Pairs of geometrically matched images: optical image (left) and corresponding SAR image (right) from the MultiSenGE dataset (for SAR images, the VV polarization in the red channel, the VH polarization in the green channel).

Although both of the above classes demonstrate relatively high quality, they also have drawbacks. A common drawback is a lack of generality. This involves the need to redefine detectors of keypoints and corresponding descriptors or search for a new transformation between image modalities whenever the properties of an image (e.g., the number of channels or their spectral arrangement) change.

Another drawback of existing approaches of the second class is the need to reconstruct specific features of one modality from images of another modality. For example, reconstructing speckle noise inherent in SAR images from optical images, or reconstructing color characteristics from SAR images.

In this paper, we propose a different approach that also utilizes transformations of input optical and SAR images. The idea behind the proposed solution is to find some "common" content for the optical and SAR images. This means finding an image that maximally preserves the common information of both images, eliminating information about the differences. In a sense, we find the "intersection" between the optical and SAR images. This approach does not require transformation of the optical image to the SAR image or vice versa, thus eliminating the second of the aforementioned drawbacks.

Images obtained by transforming optical and radar images are referred to hereinafter as *images of a shared modality* or *modality-shared images*. Since the final format (number of channels) of such images is specified a priori, their use for image co-registration proves convenient and versatile. Specifically, we can either use an existing image matching technique (applicable to images with the same number of channels) or train such a technique on new data, i.e., a set of modality-shared images extracted from the original optical and SAR images.

In this paper, we use the RoMa dense matching model [22] for image matching. As our experiments show, it can be used both fully pre-trained (on conventional images) and retrained on modality-shared images without significant changes in quality indicators. Therefore, to process images of new unknown modalities, it is not necessary to re-train the feature matching method, but it is sufficient to construct transformations into the shared modality.

The main contribution of the paper is as follows.

1. We proposed to match digital images of different modalities (obtained from different sensors) by transforming them into a shared modality. Transformed images of the same territory in this modality are similar, and the "diversity" of the original data is largely preserved.
2. We proposed a trainable technique for transforming original images into the shared modality. This technique is based on the joint training of several neural networks in the self-learning mode. It uses pairs of co-registered images of original modalities (in this study, optical and SAR images) and does not require desirable outputs, i.e., modality-shared images.
3. We trained the RoMa image matching model using modality-shared images obtained from optical and SAR images from an open dataset.
4. We have experimentally proven the advantage of the proposed approach in terms of image matching quality. Along with improved performance, the proposed solution also proves to be more versatile. Specifically, it allows the use of the existing RoMa and DeDoDe image matching models, trained on regular images, without retraining on modality-shared images, while maintaining high matching quality.

The paper is structured as follows. The second section presents a brief classification of existing co-registration techniques for optical and SAR images. This section also provides information on available datasets for the problem under consideration. The third section presents a detailed description of the proposed technique, including the overall solution pipeline and neural network architectures. The fourth section describes training details and the results of experiments. In particular, we evaluate the effectiveness of the proposed technique for optical and SAR images, and compare our solution to existing state-of-the-art techniques. The paper ends with a conclusion and a list of references. Our code is published at https://github.com/BorisovAN/shmod.

## 2 Related Work

Although geometrical co-registration of remote sensing images can be considered as a special case of image co-registration [1, 2], matching of images with different modalities (for example, optical and SAR images) requires the development of both modified and completely new solutions. Below, we provide a brief overview of approaches to optical and SAR image matching based on the classification adopted in recent relevant papers [2, 11, 23, 24].

## 2.1 Area-based Methods

Area-based methods (template-based methods) perform image matching in the process of solving an optimization problem. The optimization criterion is an aggregated value consisting of discrepancies (or similarities) calculated for pairs of corresponding image fragments. Methods of this class include, for example, mutual information-based image registration techniques [25, 26] and an improved normalized cross correlation algorithm [27].

Modern methods of this class transform image fragments into feature vectors (descriptors) and estimate the discrepancy between fragments based on the difference in their descriptors. Examples of this approach are HOG [28-30], HOPC [31], CFOG [32, 33]. An obvious disadvantage of Area-based methods (with a naive implementation) is the high computational complexity due to the comparison of a large number of fragments in images.

## 2.2 Feature-based Methods

Feature-based methods are a further development of area-based methods, eliminating their main drawback. A significant reduction in complexity is achieved by reducing the number of candidate fragments. Only fragments containing keypoints are used as candidates. Keypoints can be stably detected under various transformations, both geometric and brightness. Examples of keypoints include corners, specific boundaries, centers of closed contours, etc.

Typically, image matching using keypoints consists of three stages. First, a *keypoint detection algorithm*, called a detector, finds sets of keypoints in each image. Second, a *keypoint description algorithm* associates a numerical vector with each detected keypoint. This vector, called a descriptor, describes the properties of the keypoint's local neighborhood in the image. Third, the *keypoint matching algorithm* compares descriptors and generates pairs of corresponding points in two images being matched. In the simplest case, points are considered corresponding if the distance (Euclidean, cosine, or other) between their descriptors does not exceed a predefined threshold. The resulting geometric transformation between the images is usually described by a projective transformation matrix, which is estimated from a set of pairs of corresponding points using algorithms like RANSAC [34]. SIFT [35] is the most well-known feature-based technique for matching regular images.

The described keypoint-based approach can be effectively applied to register optical and SAR images. However, due to the different modalities of such images, it is necessary to modify the keypoint detection and description algorithms to reflect the properties of optical and SAR images. Examples of feature-based co-registration techniques are OS-SIFT [36] and PSO-SIFT [37]. The RIFT [38] and RIFT2 [39] methods, developed for constructing radiation-invariant feature transforms, also belong to the considered category. It is worth noting that the techniques listed above impose limitations on the number of image channels supporting grayscale or three-channel images. This makes these techniques unsuitable for use with multi- or hyperspectral imagery.

## 2.3 Deep Learning-based Methods

The intensive development of deep learning methods over the past decade has had a significant impact on image co-registration techniques. Neural network algorithms trained on images of specific modalities have often proven to be more effective than classical deterministic algorithms developed by researchers based on their experience. The use of deep learning allowed not only to adapt detectors and descriptors to specific observation conditions and modalities, but also to bring descriptors closer together in feature space for matching heterogeneous remote sensing images. Methods for matching homogeneous images are used in [40, 41]. Examples of methods for matching heterogeneous remote sensing images are ACAMatch [7], MatchosNet [8], CAMM-net [41], and DF-Net [42].

## 2.4 Image Translation-based Methods

Further development of deep learning techniques in remote sensing image processing has made it possible to transform images from one modality to another. For example, matching of optical and SAR images can be accomplished by transforming the optical image to a radar-like image, allowing the same detectors and descriptors to be used for both images. Such methods are often referred to as image translation-based techniques.

These transformations were made possible by generative artificial intelligence methods, particularly generative adversarial networks (GANs) [17, 18, 43, 44]. An example is the paper [45], where an optical image is translated into a SAR-like image, followed by training the MatchosNet network to match the original and generated radar images. In the paper [19], the authors investigate the quality of matching of real and generated (by converting SAR images into optical-like images) images using the SIFT, SAR-SIFT [46], and PSO-SIFT [37] techniques. An interesting approach is used in [47], where a modified CycleGAN is trained such that the information in the final latent layers of both generators of the network is similar. The information from the latent layers is then used as feature descriptors.

Overall, the approach of transforming one of the remote sensing images into an alternative modality has proven to be quite effective and is frequently used recently for object detection [48, 49] and change detection [50-52].

However, this approach has an objective drawback, resulting from the generation of a simulated image instead of a physically based image in an alternative modality. As a result, converting multi- and hyperspectral images to SAR images results in significant data loss, as these images have a significantly higher number of bands than SAR images. The inverse transformation, on the contrary, leads to a significant lack of data.

Unlike the approach based on transformation from one modality to another, the approach proposed in this paper is based on constructing an intermediate modality. This approach does not require the neural network to reconstruct details missing in the source modality; that is, the neural network does not need to "invent" details. As alternatives to the proposed approach, we consider the well-known pix2pix [16] and LSGAN [44] methods, which perform transformation from SAR to optical modality.

## 2.5 Available datasets

The following data sets are intended for conducting research on the matching of optical and SAR images.

- SEN1-2 [53] – this dataset contains 282 384 pairs of co-registered image fragments acquired by the European Sentinel-1 (SAR platform) and Sentinel-2 (multispectral sensor) satellites across the globe. This dataset uses only the VV polarization of Sentinel-1 images and only the RGB channels of Sentinel-2. The images are stored in PNG format and have a size of 256×256 pixels. The range of integer values is [0, 255] for each channel.
- SEN12-MS [54] – the dataset that differs from the SEN1-2 dataset in the data formats provided. Sentinel-1 SAR images are represented in both polarizations, while Sentinel-2 images are represented in all 13 spectral bands. This dataset also includes surface classification masks derived from MODIS data. The total dataset size is 180 662 triplets. Images are stored in TIFF format with an image size of 256×256 pixels. The range of integer values is [0, 65 535] for optical images, while SAR data is stored as real numbers.
- OsDataset [55] – the dataset from the Chinese GaoFen-3 SAR satellite, comprising 2 673 images paired with optical RGB images from Google Earth. The SAR images are presented in grayscale. The images are stored in PNG format, and their size is 512×512 pixels. The range of integer values is [0, 255].
- MultiSenGE [56] – this is a multi-temporal dataset comprising co-registered fragments of Sentinel-1 and Sentinel-2 images acquired from July to November 2020 over eastern France. The dataset also includes high-resolution land surface segmentation masks. Unlike the SEN12-MS dataset, the digital images have been pre-processed: speckle noise suppression and linearization of the SAR images have been performed; low-resolution channels 1, 9, and 10 have been excluded from the optical images (the final number of channels is 10). The dataset contains 72 033 fragments of Sentinel-2 and 1 012 227 fragments of Sentinel-1 images, corresponding to 8 157 different locations. The images are stored in TIFF format, and their size is 256x256 pixels. The range of integer values is [0, 65 535] for optical images, and the radar data is stored as real numbers.

We choose the MultiSenGE dataset for our experiments due to its multi-temporal nature. Furthermore, the radar images in this dataset do not require additional preprocessing.

## 3 Proposed Method

The general structure of the proposed method is shown in Figure 2. The method receives two images of sizes $W_1 \times H_1$ and $W_2 \times H_2$ with $K_1$ and $K_2$ channels, respectively. The method transforms each of these images into a new image with a given number of channels $K$. Unlike existing techniques based on image translation, we do not transform an image from one modality to another. Instead, we transform both images to a new modality with a given number of channels $K$, preserving the original size of the images. In this new shared modality, the method preserves the "common" content of original images and eliminates the differences between them.

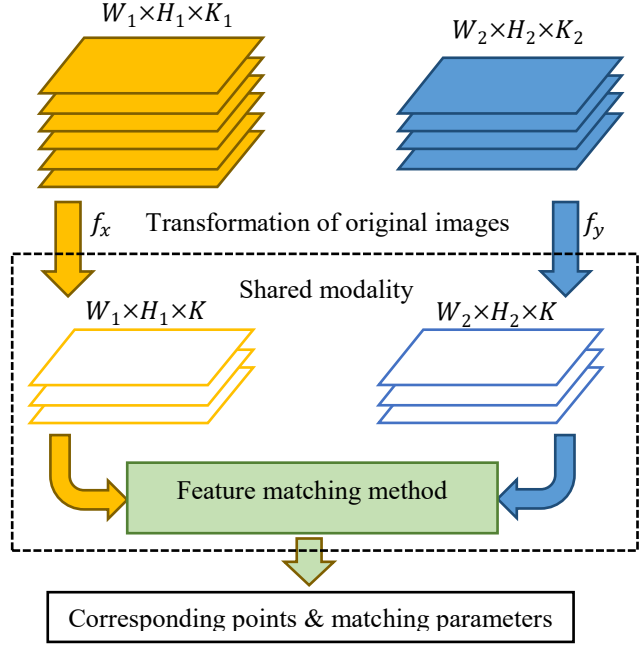

Figure 2: General scheme of the proposed method.

Below, we provide a more detailed description of the proposed method. In particular, in Section 3.1, we introduce the concept of shared modality and develop corresponding requirements. In Section 3.2, we formally pose the problem of constructing a transformation to shared modality and propose various solution options. In Section 3.3, we discuss a neural network implementation of the transformation to shared modality, and in Section 3.4, we describe the solutions to the image matching problem used in the proposed method.

## 3.1 Shared Modality

By a pair of *modality-shared images* synthesized from two corresponding images of different modalities (in particular, optical and SAR), we mean a pair of digital images that satisfy the following requirements:

− the linear dimensions $H, W$ of the modality-shared images coincide with the linear dimensions of the corresponding original images;
− modality-shared images have a predetermined and identical number of channels (hereinafter, we denote it as $K$);
− modality-shared images synthesized for matched images of different modalities should be similar (close in the sense of some measure);
− modality-shared images (and a pair of corresponding transformations) are not degenerate, that is, they retain significant information of the original images.

The last two requirements are not completely defined and are also somewhat contradictory. In particular, the third requirement (that modality-shared images should be similar) is easily satisfied by constructing degenerate transformations that synthesize constant images. The fourth requirement is designed to avoid such cases.

A possible specification of the last requirement is an accurate reconstruction of the original images from the corresponding modality-shared images. Below, we will consider this specification of the last requirement, as well as a simpler alternative. A specification of the third requirement, as well as the architecture of the constructed transformation, is also described below.

## 3.2 Transformation to Shared Modality

### 3.2.1 Construction of the Transformation

Let $S = \{(x_i, y_i) | x_i \in \mathbb{R}^{W \times H \times K_X}, y_i \in \mathbb{R}^{W \times H \times K_Y}\}_{i=1}^N$ be a set of matched pairs of digital remote sensing images of size $W \times H$ with $K_X$ and $K_Y$ channels, respectively. For the sake of certainty, we will assume that images $x_i$ are SAR

images of modality $X$, and $y_i$ are optical images of modality $Y$. *To transform original images into a shared modality*, we solve the optimization problem of finding a pair of mappings $(F_X, F_Y)$ of the following form:

$$(F_X, F_Y) = \underset{\substack{f_X \in \mathcal{F}_X, \\ f_Y \in \mathcal{F}_Y}}{\operatorname{argmin}} \Phi(f_X, f_Y),$$

$$\Phi(f_1, f_2) = \mathbb{E}_{(x,y) \in S}\left[L\big(f_X(x), f_Y(y)\big) + \alpha D\big(x, f_X(x), y, f_Y(y)\big)\right]. \tag{1}$$

Here, $L: \mathbb{R}^{W \times H \times K} \times \mathbb{R}^{W \times H \times K} \to \mathbb{R}$ is a function that evaluates the similarity of two modality-shared images; $D$ is a real-valued function of four arguments $\mathbb{R}^{W \times H \times K}$, estimating the degeneracy of modality-shared images $f_X(x)$ and $f_Y(y)$ relative to original images $x$ and $y$; $\mathcal{F}_X$ and $\mathcal{F}_Y$ are the classes of the transformations $f_X$ and $f_Y$ having the form $f_*: \mathbb{R}^{W \times H \times K_*} \to \mathbb{R}^{W \times H \times K}$, for which the solution is searched; $\alpha \in \mathbb{R}_+$ is a weighting coefficient.

Note that the proposed approach differs significantly from existing image translation-based solutions (see Section 2.4) in the following two aspects.

*First*, when searching for a transformation to an existing modality, the desired transformation result (an image of the alternative modality) already exists and can be used for training. When searching for a transformation to a shared modality, the desired result is unknown, and the "true" output of the DNN and the final transformation are determined through training. Thus, the result is largely determined by the criteria $L$ and $D$, the specified transformation classes $\mathcal{F}_*$ (i.e., the type of DNNs), and the training techniques.

*Second*, existing solutions based on image transformations between original modalities require reconstructing specific image features in these modalities. This includes reconstructing speckle noise inherent in SAR images from optical images or reconstructing color characteristics from SAR image data. In the proposed approach, such reconstruction of properties unnatural for the original images is not required. Clearly, this is a positive feature allowing for improved image registration quality (we demonstrate it in Section 3).

We already considered a similar optimization problem for spectral matching of optical images in the paper [57]. We have demonstrated that a sufficiently good (linear) solution can be found analytically, and the best result can be achieved by utilizing neural networks to implement transformations $\mathcal{F}_*$.

### 3.2.2  Criteria: similarity measure

The requirement for "non-degeneracy" of modality-shared images means that images do not degenerate into (almost) constant regions. On the other hand, the requirement for similarity of modality-shared images means that such images must share a common set of spatial features, and the average brightness must be the same. Consequently, when constructing the function $L$, both differences in image brightness and the preservation of spatial features must be taken into account.

In the paper [58], various loss functions are considered for image reconstruction problems. It is shown that adding the Structural Similarity Index Measure (SSIM) [59] to the loss function allows for better results compared to using only Mean Squared Error (MSE) or Mean Absolute Error (MAE).

Therefore, we use the following function $L$ to define the similarity measure of two modality-shared images $f_X(x)$ and $f_Y(y)$:

$$L\big(f_X(x), f_Y(y)\big) = \beta MSE\big(f_X(x), f_Y(y)\big) + \gamma DiSSIM\big(f_X(x), f_Y(y)\big). \tag{2}$$

Here $\beta, \gamma \in \mathbb{R}_+$ are weighting coefficients, $MSE$ is the MSE between images, and the $DiSSIM$ function is defined using SSIM as follows:

$$DiSSIM\big(f_X(x), f_Y(y)\big) = 1 - SSIM\big(f_X(x), f_Y(y)\big),$$

where

$$SSIM(A, B) = \frac{(2E[A]E[B] + c_1)(2S[AB] + c_2)}{(E^2[A] + E^2[B] + c_1)(S^2[A] + S^2[B] + c_2)}.$$

Here, $A, B \in \mathbb{R}^{w \times w}$ are grayscale images, $E[A]$ and $E[B]$ are mean values, $S^2[A]$ and $S^2[B]$ are sample variances, $S[AB]$ is the covariance of images $A$ and $B$, and $c_1$ and $c_2$ are regularization constants. The value of SSIM belongs to the range [-1.0, 1.0], where 1.0 means the highest image similarity. For images larger than $w \times w$, the output value is the average SSIM value for all windows of size $w \times w$. If processed images are not grayscale, then the average SSIM value across all channels is used. In this paper, we set $w$ to 5 to better preserve small details.

It is worth noting that the paper [58] also considered the use of Multiscale SSIM (MS-SSIM) and MAE, but experiments showed that the best results were obtained with a combination of MSE and SSIM.

### 3.2.3 Criteria: degeneracy measure

Let's consider the measure $D(\cdot)$, which evaluates the degeneracy of modality-shared images $f_X(x)$ and $f_Y(y)$ relative to the corresponding original images $x$ and $y$. In this paper, we consider two versions of this measure, which differ significantly in both architecture and computational costs.

The *first version* directly implements the last requirement for modality-shared images (See Section 3.1) as a requirement for high-quality reconstruction of original images from modality-shared images. In this case, the problem (1) can be viewed as the process of training a pair of autoencoders [60] with an additional constraint on the closeness of their latent representations. Let $f_X(x)$ and $f_Y(y)$ be encoders, and the decoders implement the inverse transforms $f_X^{-1}(\cdot)$ and $f_Y^{-1}(\cdot)$. Then $D$ can be defined as a weighted sum of two losses, common for autoencoder training:

$$D(x, f_X(x), y, f_Y(y)) = L_D\left(x, f_X^{-1}(f_X(x))\right) + \eta L_D\left(y, f_Y^{-1}(f_Y(y))\right), \tag{3}$$

$$L_D(a, b) = MSE(a, b) + DiSSIM(a, b)$$

Here, $L_D: \mathbb{R}^{W \times H \times K} \times \mathbb{R}^{W \times H \times K} \to \mathbb{R}$ is a function characterizing the measure of similarity between two images of a given modality, and $\eta \in \mathbb{R}_+$ is a weighting coefficient that balances the values of this function for different modalities.

Reconstruction of original images from modality-shared images is complex both architecturally and computationally, so we consider the simpler second version of the degeneracy measure $D$ below.

The *second version* of the $D$ measure exploits the similarity of grayscale images to ensure non-degeneracy of the transformations. In the case of a SAR source image, the grayscale image is extracted from the corresponding modality-shared image. In the case of an optical source image, the grayscale image is extracted from its RGB components (or other components with the highest resolution). Thus, the second version of the $D$ measure is as follows:

$$D(x, f_X(x), y, f_Y(y)) = DiSSIM\left(g(f_X(x)), g(y|_{RGB})\right). \tag{4}$$

Here, $g: \mathbb{R}^{W \times H \times \cdot} \to \mathbb{R}^{W \times H \times 1}$ is a function that averages image channels, and $y|_{RGB}$ "selects" RGB channels from a (multi-) spectral optical image (it is trivial for ordinary RGB images).

The second version of the $D$ measure eliminates the need to construct inverse transformations for both original modalities and requires a smaller number of trained DNNs in general.

It is worth noting that the $D$ measure was initially considered as:

$$D(x, f_X(x), y, f_Y(y)) = \lambda_1 DiSSIM\left(g(f_X(x)), g(y|_{RGB})\right) + \lambda_2 DiSSIM\left(g(f_Y(y)), g(y|_{RGB})\right).$$

However, values of $\lambda_2 > 0$ resulted in deterioration of quality indicators. We can explain this based on the following considerations:

- If $\lambda_2 = 0$, the value of (1) depends only on the similarity of modality-shared images and the structural similarity of the SAR image in shared modality and the optical grayscale image. We can say that $f_X(x)$ is trained to reconstruct the structural features of an optical image using SAR. Since complete reconstruction is impossible and some details are lost, $f_Y(y)$ is trained to simplify the optical image, losing those details that cannot be reconstructed by $f_X(x)$.
- If $\lambda_2 > 0$, any loss of detail in the optical image increases the value of (1), meaning $f_Y(y)$ will discard less detail, which will increase the difference in the outputs of $f_X(x)$ and $f_Y(y)$.

Figure 3 shows both considered versions of the measure $D$, including the DNN training architecture. The following notations are used in the figure: $x$ and $y$ are co-registered SAR and optical images; $\tilde{x} = f_X(x)$ and $\tilde{y} = f_Y(y)$ are corresponding modality-shared images; $\hat{x} = f_X^{-1}(\tilde{x})$ and $\hat{y} = f_Y^{-1}(\tilde{y})$ are corresponding reconstructed images of the original modalities. The DNNs being trained are marked in gray in the figure.

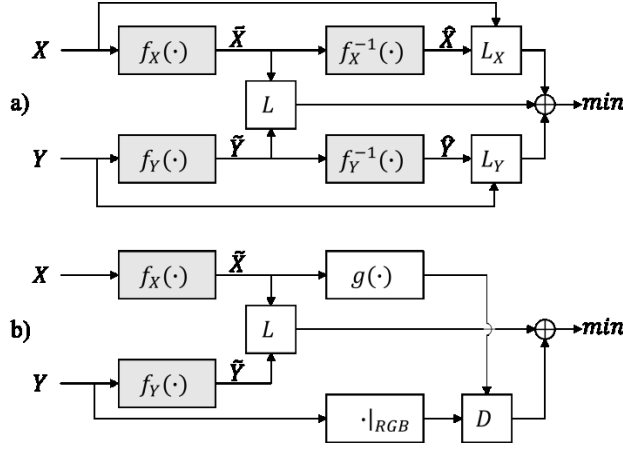

Figure 3: Training of transformations into a shared modality: pipeline for criterion (3) - based transformation (a); pipeline for criterion (4) - based transformation (b).

### 3.3 Network Architectures

To construct the transformations $\tilde{x} = f_X(x)$ and $\tilde{y} = f_Y(y)$ of the original images into shared modality, we used a U-Net [61] – based framework. In particular, we used the Efficient-Net-B0 network [62] as the encoding part to improve the generated descriptions. We set the number of layers in this network to 5. We placed a sigmoid activation function, which mapped the output to the range (0.0, 1.0), at the end of the network.

To implement the reconstructive transformations $\hat{x} = f_X^{-1}(\tilde{x})$ and $\hat{y} = f_Y^{-1}(\tilde{y})$ from modality-shared images into original modalities, we used a similar network with an inverse number of input and output channels, but without the sigmoid function at the network output.

### 3.4 Matching of Modality-shared Images

After converting images to a shared modality, they can be co-registered using regular image matching methods. The advantage of this approach is the availability of ready-made implementations of such methods in common computer vision libraries. The most well-known method for matching regular images is SIFT [35].

Among the available state-of-the-art neural network models for image matching, the DeDoDe and RoMa models stand out.

The DeDoDe model [63] is a feature detection method. It consists of two networks: a keypoint detection network and a description network that calculates a feature description of the keypoints. Feature descriptors are compared based on their proximity.

The RoMa model [22] is a so-called dense matching method. The result of the method is a map of correspondences between all pixels of the first and second images, along with a quality score for each pair. This approach has the potential to find correspondences between images under conditions of strong nonlinear distortions.

It's worth noting that orthorectified remote sensing images are related by an affine transformation. Using a dense mapping for such images is redundant. Therefore, to evaluate the transformation parameters, we select some point pairs with the highest quality.

## 4 Experiments

### 4.1 Data

For our studies, we used the MultiSenGE dataset [56], which contains both optical and SAR images. Since MultiSenGE contains multiple SAR and optical images acquired at different times for each region, we formed training pairs as follows: for each optical image, we selected a radar image for the same region with the closest acquisition date. The resulting dataset was divided into training, test, and validation sets to avoid territorial overlap (all pairs of images for a given region are contained entirely within one of the sets). The training set consisted of 46 092 pairs of images, the validation set consisted of 11 522 pairs, and the test set consisted of 14 405 pairs of images. The SAR image data were converted to a logarithmic scale.

At the time of the experiments, existing software solutions for training generative networks supported only RGB or grayscale images, but not multispectral ones. For this reason, we created auxiliary sets of grayscale and RGB images from the original data as follows:

- we converted radar images into grayscale images containing only VV polarization in dB scale (similar to the SEN1-2 set);

- we converted optical (multispectral) images into color images with RGB channels.

To convert the range of image pixel values to a single interval [0, 1], we used the following procedure. For each image modality (optical and SAR), we calculated the global percentiles $p_{0.5}$ and $p_{99.5}$, and constrained the range of values to the interval $[p_{0.5}, p_{99.5}]$ with subsequent linear transformation to the range [0, 1].

### 4.2 Equipment and Software

The experimental studies were conducted using a computer equipped with an AMD Ryzen 5600X CPU, 32 GB of RAM, and an Nvidia GeForce RTX 3090 GPU.

We used PyTorch 2.7 [64] as a machine learning framework. We also used the U-Net implementation with various encoders from the segmentation-models-pytorch package [65].

### 4.3 Alternative Techniques and Training Details

In our study, we compared the proposed method for optical and SAR image matching with modern techniques based on Image Translation and the best feature matching algorithms.

We used the well-known generative models LSGAN [44] and pix2pix [16] as Image Translation-based methods. The implementation of the networks was taken from a public repository [66]. We trained the pix2pix GAN models with default parameters without changing the code, and trained LSGAN with identity loss disabled. To improve the convergence of LSGAN, we set the learning rate of the generator and discriminator to 0.0002 and 0.001, respectively, according to the Two Time-Scale Update Rule (TTUR) [67]. Without applying TTUR, the LSGAN method showed unsatisfactory results.

We considered four feature selection/matching techniques to co-register images of original modalities. Specifically, the techniques were SIFT [35], RIFT2 [39], DeDoDe [63], and RoMa [22]. Furthermore, we used the same four techniques as possible components of the proposed approach to match modality-shared images. To the best of our knowledge, the DeDoDe and RoMa methods *have not been applied to remote sensing data* by other authors to date (they have been used only for matching conventional images).

Since matching techniques based on neural networks can be adapted to a specific modality, we decided to train the feature-matching network of the best-performing combination (the proposed method based on criterion (4) in combination with the RoMa model) on modality-shared images. Thus, we trained the RoMa neural network on modality-shared images to improve co-registration performance. In the training process, we used augmentation with a random translation up to 64 pixels, random rotation up to 30°, and random scaling with a coefficient in the range [0.9, 1.1]. We adhered to standard parameters, except for the image size, which was set to 224×224.

### 4.4 Training of Transformation to Shared Modality

For all variants of transformation into shared modality, we used a final number of channels of $K = 3$. This allowed us to apply a wide range of feature matching techniques.

The parameters used to train the neural networks in the proposed method are listed below.

*General parameters:*

- $K_1 = 2, K_2 = 10$ (Eq. 1);
- optimizer: NAdam [68];
- base learning rate: $2 \times 10^{-4}$;

- number of epochs: 512;
- batch size: 4;
- batches per epoch: 1024;
- warmup learning reate: $10^{-8}$;
- warmup epochs: 4;
- warmup learning rate scheduling: linear;
- augmentations: random flips and 90° rotations.

*Criteria - version 1*

- $\alpha = 1$ (Eq. 1);
- $\beta = 3$ (Eq. 2);
- $\gamma = 16$ (Eq. 2);
- $\eta = 1$ (Eq. 3);

*Criteria - version 2*

- $\alpha = 16$ (Eq. 1);
- $\beta = 1$ (Eq. 2).

## 4.5 Quality indicators

The quality of remote sensing image matching was assessed by finding the matrix $T \in \mathbb{R}^{3\times3}$ of the projective transformation between optical and SAR images. We calculated the matrix of the desired projective transformation from a set $P = \{(p_1, p_2)_k | p_1 \in P_X, p_2 \in P_Y\}$ of pairs of corresponding points. Here, $P_X = \{p^i \in \mathbb{R}^2\}$ is the set of keypoints in the SAR image, and $P_Y$ is the set of keypoints in the optical image, found by the proposed or alternative methods.

To find the corresponding keypoints, we employed the RANSAC algorithm [34] with a maximum number of iterations set to 2000. Note that RANSAC allows us to find not only the estimated matrix $\hat{T}$ of the desired projective transformation, but also:

- operator $t(p, T)$ of the projection of a point from one image to a point in another image;

- a subset $P_+ = \{(p_1, p_2) \in P | \|t(p_1, \hat{T}) - p_2\| \leq 3\}$ of pairs of points consistent with the matrix $\hat{T}$ (i.e., pairs of points, the distance between which, after projection, does not exceed the predefined threshold of 3 pixels).

Using the notations introduced above, we determine the following image matching quality indicators.

The *Mean Matching Accuracy (MMA)* is defined as the fraction of corresponding points consistent with the evaluated projective transformation:

$$MMA = \frac{|P_+|}{|P|}.$$

The *Average Corner Error (ACE)* is equal to the average misalignment of the corner points of the optical and SAR images after their co-registration:

$$Corners = \{(0,0), (0,H), (W,0), (W,H)\},$$

$$ACE = \frac{1}{4} \sum_{c \in Corners} \|t(c, \hat{T}) - c\|.$$

Here $\|\cdot\|$ is the Euclidean norm.

The *Success rate (SR)* is defined as the proportion of images with ACE below a given threshold: $ACE < 40$ (in case of matching "failure", *ACE* is considered equal to infinity).

Thus, the *MMA* and *SR* indicators characterize the proportion of correct matches. The indicators described below show coordinate matching errors.

The $\Delta_P$ *indicator* is defined as the average distance between the corresponding points (note that from here on, the true transformation $T$ is used, not the estimated one):

$$\Delta_P = \frac{1}{|P|} \sum_{(p_1, p_2) \in P} \|t(p_1, T) - p_2\|.$$

To determine subsequent indicators, we introduce the concept of a *correct pair*, as a pair of keypoints whose distance does not exceed a given threshold δ. We denote the set of correct pairs by:

$$P_\delta^{Correct} = \{(p_1, p_2) \in P \mid \|t(p_1, T) - p_2\| \leq \delta\}.$$

The *Correct Matches Rate (CMR)* indicator is defined as the ratio of the number of correct pairs to their total number:

$$CMR_\delta = \frac{|P_\delta^{Correct}|}{|P|}.$$

The *Localization Error (LE)* indicator is defined as the average distance between points in a correct pair:

$$LE_\delta = \frac{1}{|P_\delta^{Correct}|} \sum_{(p_1, p_2) \in P_\delta^{Correct}} \|t(p_1, T) - p_2\|.$$

Along with *SSIM*, the following indicators are used to measure the similarity of modality-shared images $\tilde{x}$ and $\tilde{y}$:

$$RMSE(\tilde{x}, \tilde{y}) = \sqrt{MSE(\tilde{x}, \tilde{y})};$$
$$PSNR(\tilde{x}, \tilde{y}) = 10 \lg \frac{MAX^2}{MSE(\tilde{x}, \tilde{y})},$$

where *MAX* is the maximum possible pixel value (in our case, *MAX*=1.0).

## 4.6 Results

At the first stage of our experiments, we examined the similarity between matched images obtained using different transformation techniques. To do this, we took pairs of aligned original optical and SAR images and compared these images and their transformations into alternative and shared modalities.

Some visual examples of this analysis are shown in Figure 4. In this figure, the first two rows show aligned pairs of optical and SAR images. The next two rows (III and IV) show the results of transforming the SAR images into optical modality using the Pix2pix and LSGAN methods, respectively. The last two pairs of rows (V-VI and VII-VIII) demonstrate the results of transforming optical and SAR images into a shared modality for criteria (3) and (4), corresponding to the first and second transformation options.

Clearly, the results of these transformations differ significantly from each other. It should also be noted that the modality-shared images obtained using criterion (3) have low contrast. Therefore, we applied the normalization described in Section 4.1 before calculating the quality indicators. The modality-shared images obtained using criterion (4) have sufficient dynamic range and do not require normalization.

The quantitative results of the analysis are shown in Table 1. In this table, the rows correspond to different transformation techniques. In particular, the first row (labeled as "None") corresponds to matching original images without any modality transformations. In this case, to match optical and radar images, we used a color RGB image extracted from the optical image and a three-channel image obtained by repeating the VV channel of the SAR image. A similar approach was used in the SEN1-2 dataset. We normalized the image data to the range [0, 1] using the procedure described in Section 4.1 for analysis.

Table 1: Similarity measures for pairs of images obtained using different transformation techniques.

| Method | RMSE ↓ | PSNR ↑ | SSIM ↑ |
|---|---|---|---|
| None | 0.41 | 7.92 | 0.04 |
| Pix2pix | 0.16 | 16.78 | 0.35 |
| LSGAN | 0.19 | 15.30 | 0.32 |
| Ours-1 | *0.11* | *19.55* | *0.46* |
| Ours-2 | **0.07** | **23.9** | **0.78** |

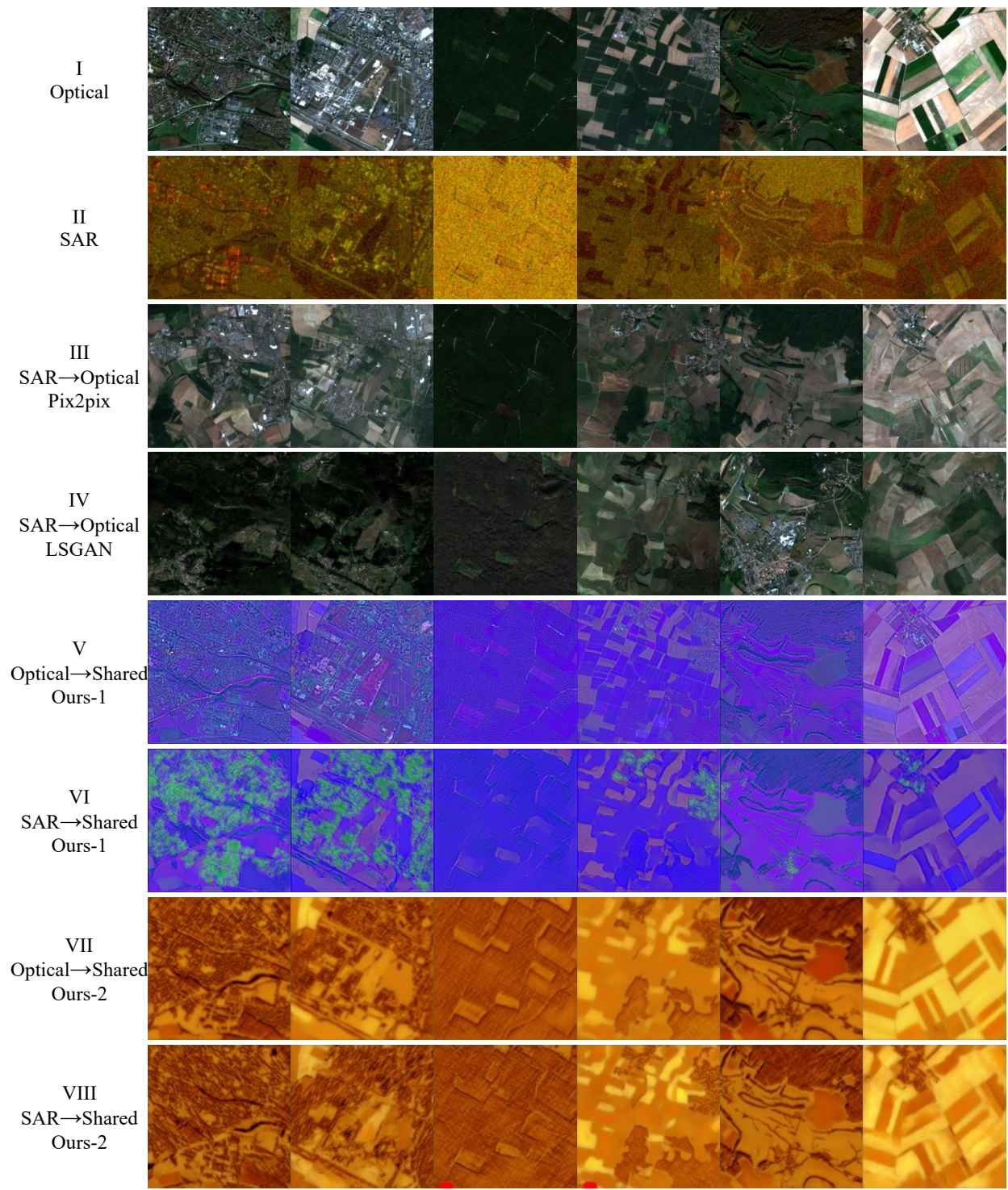

Figure 4: Examples of original images ("Optical" and "SAR") and their transformations into alternative ("Optical ->SAR" and "SAR->Optical") and shared ("Optical->Shared" and "SAR->Shared") modalities.

The remaining rows correspond to alternative techniques (Pix2pix and LSGAN) based on translation to an alternative modality, and the proposed technique in two variants ("Ours-1" and "Ours-2") described above.

The considered Table 1 shows the values of three similarity indicators, namely, *RMSE*, *PSNR*, and *SSIM*. Arrows next to the column headings indicate which values are better. For *RMSE*, lower values are better, while for the *PSNR* and *SSIM* indicators, higher values are better. The best values for the corresponding criterion are shown in bold, while values closest to the best are shown in italics.

It is clear from the considered table that the similarity between optical and SAR images in the shared modality is significantly higher (by several times) than between optical and pseudo-optical images obtained from SAR images using Pix2pix and LSGAN transformations.

In the second stage of experiments, we performed a numerical comparison of the proposed method with alternative techniques based on the criteria outlined in Section 4.5. The results of the experiments are shown in Table 2. As before, the list of alternative techniques includes the direct matching of an optical RGB image to a grayscale radar image (denoted as "None"), the Pix2pix, and LSGAN techniques. The proposed technique is presented in two versions ("Ours-1" and "Ours-2"), corresponding to different versions of the degeneracy measure (see equations (3) and (4)). For each transformation technique considered, we show the results for all four feature matching methods: SIFT, DeDoDe, RIFT2, and RoMa.

For each combination of transformation technique and matching method, we provide the numerical value of the six quality indicators described in the previous section. In addition, we provide the average number of pairs of corresponding points $|P|$ and the number of correct pairs $|P_5^{Correct}|$. Arrows in the corresponding column headings indicate whether lower or higher values are better. Specifically, for *SR*, *MMA*, and *CMR*, higher values are better. For *ACE*, $\Delta_P$, and *LE*, lower values are better.

In the considered table, the best criterion values are shown in bold, the second-best values are shown in italics; underlined values correspond to the best feature selection method for a given modality transformation technique.

It is worth noting that in this experiment, we limited the number of detected keypoints to 4096. Furthermore, when calculating the average *ACE* and *MMA* indicators, we only considered successfully matched images.

Note also that most of the small $LE_5$ values (0.09, 0.99, 0.41, 0.63) for the SIFT matching method are due to the almost complete absence of correct pairs of corresponding points ($|P_5^{Correct}|<1$). For this reason, we disregard $LE_5$ for these cases and show such values in square brackets in the table.

Clearly, the proposed technique, transforming optical and SAR images into a shared modality, demonstrates the best quality indicators. The best result was achieved by the second version of the proposed method ("Ours-2"), corresponding to criterion (4), for feature selection using the RoMa network trained on modality-shared images.

It is important that the second-best results were achieved by the second version of the proposed method ("Ours-2") with the RoMa and DeDoDe models pre-trained on regular images (i.e., without training the models on modality-shared images). In particular, the proposed method in combination with RoMa ranked second in *ACE*, *MMA*, *CMR*, and *LE*. In combination with DeDoDe, it was second in the indicators *SR* and $\Delta_P$.

This result demonstrates the universality of the proposed approach. It can be assumed that for matching images of other modalities (different from those considered in this paper), it is sufficient to train a transformation to a shared modality and use the pre-trained RoMa or DeDoDe models for feature selection without significant loss of image co-registration quality.

To comprehensively evaluate the quality of the matching, we conducted experiments for different values of the threshold δ used to calculate the $CMR_\delta$ and $LE_\delta$ indicators. Figure 5 shows the dependence of the *CMR* and *LE* indicators on the threshold value. Based on the *SR* indicator, we present only the best combinations for each transformation technique.

We observed that both indicators increase monotonically with δ, which is expected. The proposed "Ours2+RoMa (trained)" solution takes the lead. The curve for this combination lies above the other curves for the *CMR* indicator (Figure 5a) and below the other curves for the *LE* indicator (Figure 5b). Note that the number of point pairs with a coordinate error of less than 1 pixel is more than 60% ($CMR_\delta$ = 0.63), and 98% of the pairs have a coordinate error of no more than 3 pixels.

Some examples of feature detection and matching are shown in Figure 6. In this figure, for brevity, we present only the best combinations of transformation techniques and matching methods. Specifically, we present the results of matching images in the original ("None"), alternative (Pix2pix and LSGAN), and shared modalities using the DeDoDe method. For the proposed technique based on transformation to shared modality, we present the matching results using the RoMa model trained on modality-shared images.

In the considered figure, correct matches are highlighted in green, and incorrect matches are highlighted in red. Even a visual comparison reveals that the proposed approach ("Ours-1" and "Ours-2"), based on transforming to shared modality, outperforms the approaches based on transforming images to an alternative modality.

Table 2. Image matching quality indicators obtained using different combinations of transformation techniques and feature matching methods.

| Transformation technique | Matching method | $SR \uparrow$ | $ACE \downarrow$ | $MMA \uparrow$ | $\Delta_P \downarrow$ | $|P|$ | $|P_5^{Correct}|$ | $CMR_5 \uparrow$ | $LE_5 \downarrow$ |
|---|---|---|---|---|---|---|---|---|---|
| None | SIFT | 0.0001 | 12.78 | 0.51 | 126.6 | 15.3 | 0.1 | 0.007 | [0.09] |
| | DeDoDe | 0.26 | 20.02 | 0.14 | 88.6 | 152.7 | 13.7 | 0.09 | 2.71 |
| | RIFT2 | 0.03 | 22.13 | 0.32 | 63.2 | 29.5 | 3.8 | 0.13 | 1.76 |
| | RoMa | 0.01 | 16.52 | 0.30 | 155.6 | 4096 | 12.3 | 0.003 | 2.04 |
| Pix2pix | SIFT | 0.02 | 17.5 | 0.48 | 100.5 | 5.2 | 0.7 | 0.13 | [0.99] |
| | DeDoDe | 0.93 | 7.28 | 0.26 | 26.4 | 400 | 172.0 | 0.43 | 2.65 |
| | RIFT2 | 0.008 | 25.4 | 0.56 | 26.5 | 7.7 | 2.5 | 0.33 | 1.88 |
| | RoMa | 0.28 | 13.37 | 0.23 | 99.5 | 4096 | 491.5 | 0.12 | 2.68 |
| LSGAN | SIFT | 0.01 | 14.17 | 0.50 | 112.5 | 3.9 | 0.4 | 0.09 | [0.63] |
| | DeDoDe | 0.60 | 10.26 | 0.26 | 61.2 | 229.2 | 59.6 | 0.26 | 2.42 |
| | RIFT2 | 0.014 | 24.22 | 0.53 | 33.7 | 9 | 3.0 | 0.33 | 1.88 |
| | RoMa | 0.03 | 13.69 | 0.22 | 143.8 | 4096 | 81.9 | 0.02 | 2.13 |
| Ours-1 | SIFT | 0.0001 | 14.16 | 0.78 | 97.8 | 1.2 | 0.2 | 0.17 | [0.41] |
| | DeDoDe | 0.99 | 3.01 | 0.51 | 7.4 | 553.1 | 409.3 | 0.74 | 2.06 |
| | RIFT2 | 0.31 | 17.68 | 0.31 | 35.5 | 83.9 | 25.2 | 0.3 | 2.56 |
| | RoMa | 0.69 | 3.51 | 0.6 | 41.6 | 4096 | 2007.0 | 0.49 | 1.68 |
| Ours-2 | SIFT | 0.74 | 7.13 | 0.68 | 21 | 20.1 | 15.1 | 0.75 | 1.76 |
| | DeDoDe | *0.996* | 1.96 | 0.68 | *3.6* | 877 | 771.8 | 0.88 | 1.67 |
| | RIFT2 | 0.67 | 14.3 | 0.31 | 17.3 | 17.3 | 8.0 | 0.46 | 2.77 |
| | RoMa | 0.97 | *1.56* | 0.85 | 6.7 | 4096 | 3727.4 | *0.91* | *1.12* |
| | RoMa (trained) | **0.999** | **0.85** | **0.94** | **1.0** | 4096 | 4083.7 | **0.997** | **0.95** |

*) The best values are shown in bold, the second-best values are shown in italics; the best values for a given modality transformation technique are underlined. $LE_5$ values ignored due to insufficient number of correct pairs are given in square brackets.

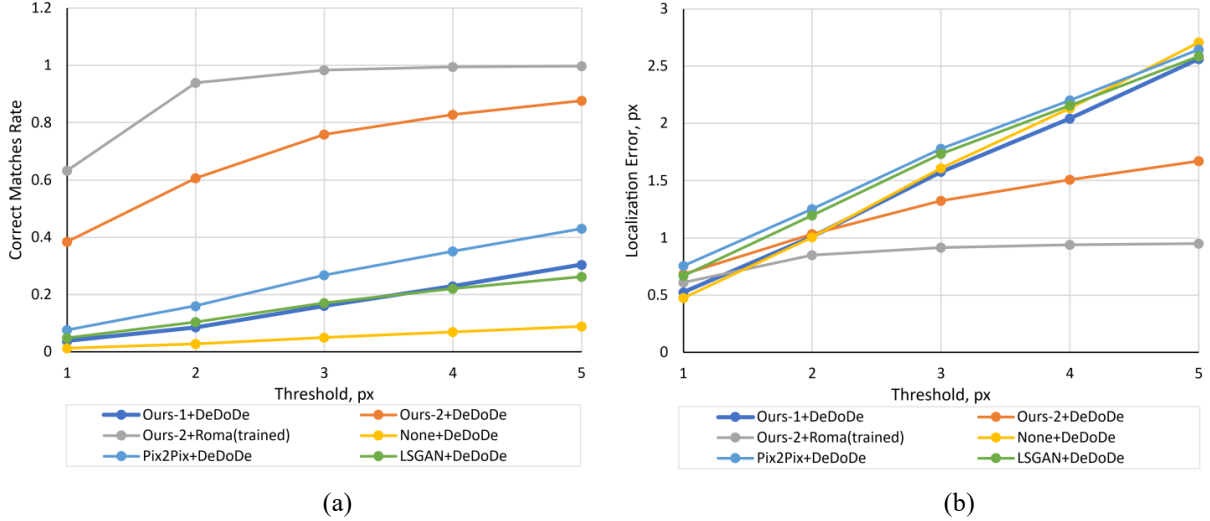

(a)                                                                                                 (b)

Figure 5: Dependence of the quality indicators on the threshold value δ: Correct Matches Rate $CMR_\delta$ (a), and Localization Error $LE_\delta$ (b).

## 5  Conclusion

In this paper, we proposed a new method for matching optical and SAR Earth remote sensing images. The key element of the method is the transformation of images from the original modalities into a new modality. Unlike existing solutions based on transforming between the modalities of original images, the proposed technique extracts a new shared modality based on the analysis of both original images. In this shared modality, it is possible to achieve both the similarity of images transformed from the original modalities and the preservation of the information required to solve the co-registration problem.

We formalized the problem of constructing the transformation to the shared modality as an optimization problem and obtained its solution for two particular criteria in the form of corresponding transformations using deep learning-based techniques.

In the final solution, we propose to use the RoMa neural network model as a feature matching technique for the co-registration of modality-shared images. To improve the matching quality, we proposed training the RoMa model on a set of modality-shared image pairs.

We conducted an experimental study on the MultiSenGE dataset, which included both optical and SAR images. The results demonstrated the superiority of the proposed solution compared to the best image translation-based solutions. Furthermore, we demonstrated the universality of the proposed approach: the quality advantage was maintained when the pre-trained RoMa and DeDoDe models were used without re-training on modality-shared images.

A promising direction for the development of the proposed approach seems to be the search for more effective components of the considered optimization problem, namely, measures of similarity and degeneracy, as well as neural network models for constructing transformations between original and shared modalities.

## Acknowledgments

This work was supported by the Ministry of Science and Higher Education of the Russian Federation, project no. 075-15-2024-558.

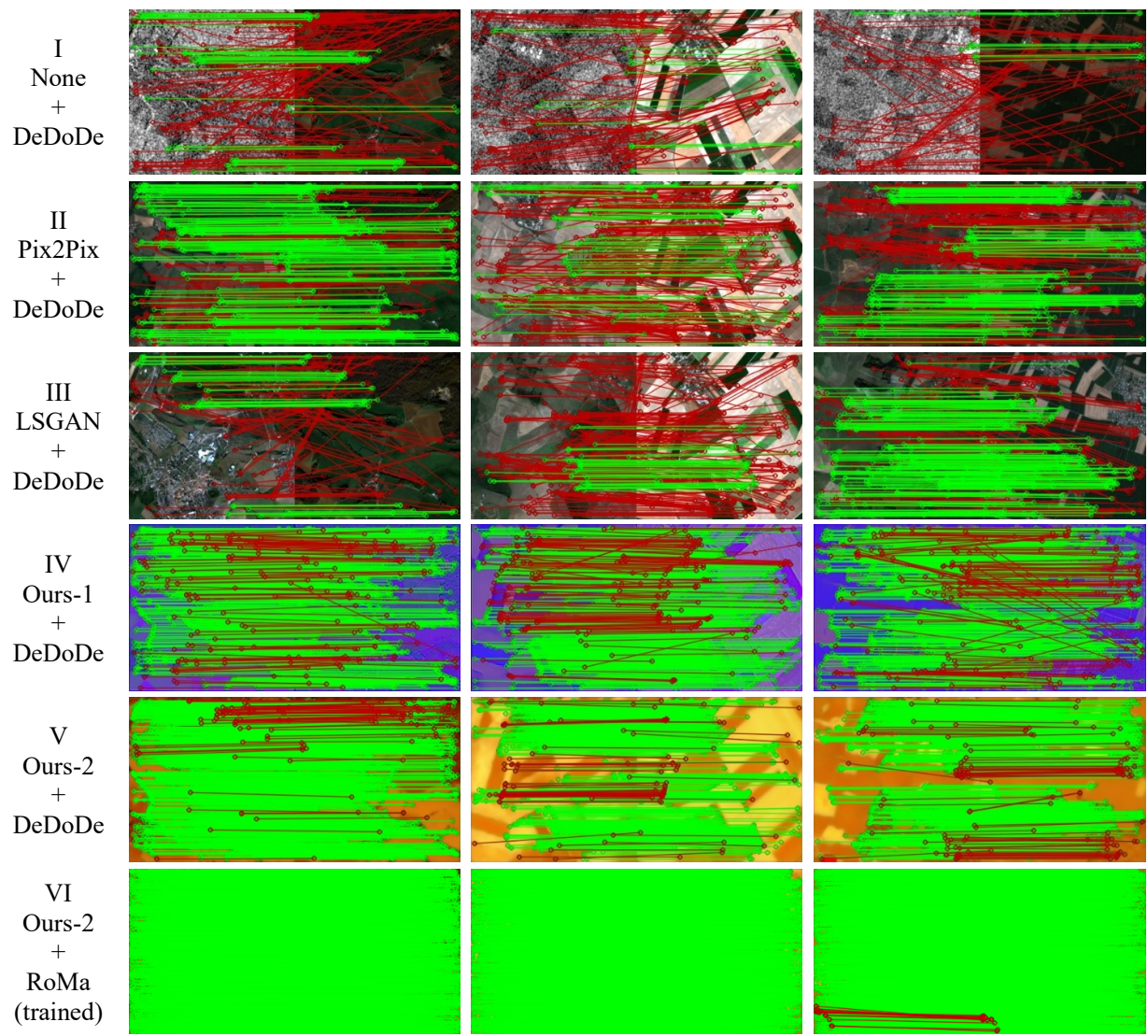

Figure 6: Examples of feature detection and matching for various combinations of transformation techniques and feature matching methods.